\def\year{2020}\relax
\documentclass[letterpaper]{article} 
\usepackage{aaai20}  
\usepackage{times}  
\usepackage{helvet} 
\usepackage{courier}  
\usepackage[hyphens]{url}  
\usepackage{graphicx} 
\usepackage{graphicx}  
\usepackage{multirow}
\usepackage{amsmath}
\usepackage{amssymb}
\usepackage{wasysym}
\usepackage{rotating}

\newcommand{\parencite}[1]{( \citeauthor{#1} \citeyear{#1})}
\providecommand{\tabularnewline}{\\}
\title{EHSOD: CAM-Guided End-to-end Hybrid-Supervised Object Detection\\ with Cascade Refinement}
\author{Linpu Fang,\textsuperscript{\rm 1 \thanks{Both authors contributed equally to this work.}} Hang Xu,\textsuperscript{\rm 2 $^{*}$}\thanks{Corresponding Author: xbjxh@live.com} Zhili Liu,\textsuperscript{\rm 2}\\ \Large \textbf{Sarah Parisot,\textsuperscript{\rm 2} Zhenguo Li\textsuperscript{\rm 2}} \\ 
	\textsuperscript{\rm 1}South China University of Technology\\ 
	\textsuperscript{\rm 2}Huawei Noah's Ark Lab}

\begin{document}
\maketitle
\begin{abstract}
Object detectors trained on fully-annotated data currently yield state
of the art performance but require expensive manual annotations. On
the other hand, weakly-supervised detectors
have much lower performance and cannot be used reliably in a realistic
setting. In this paper, we study the hybrid-supervised object detection
problem, aiming to train a high quality detector with only a limited amount of fully-annotated
data and fully exploiting cheap data with image-level labels. State
of the art methods typically propose an iterative approach, alternating
between generating pseudo-labels and updating a detector. This paradigm
requires careful manual hyper-parameter tuning for mining good pseudo
labels at each round and is quite time-consuming. To address these
issues, we present EHSOD, an end-to-end hybrid-supervised object detection
system which can be trained in one shot on both fully and weakly-annotated
data. Specifically, based on a two-stage detector, we proposed two
modules to fully utilize the information from both kinds of labels:
1) CAM-RPN module aims at finding foreground proposals guided by a
class activation heat-map; 2) hybrid-supervised cascade module further
refines the bounding-box position and classification with the help
of an auxiliary head compatible with image-level data. Extensive experiments
demonstrate the effectiveness of the proposed method and it achieves
comparable results on multiple object detection benchmarks with only
30\% fully-annotated data, e.g. 37.5\% mAP on COCO. We will release
the code and the trained models.
\end{abstract}

\section{Introduction}

\begin{figure}[th]
\begin{centering}
\includegraphics[scale=0.15]{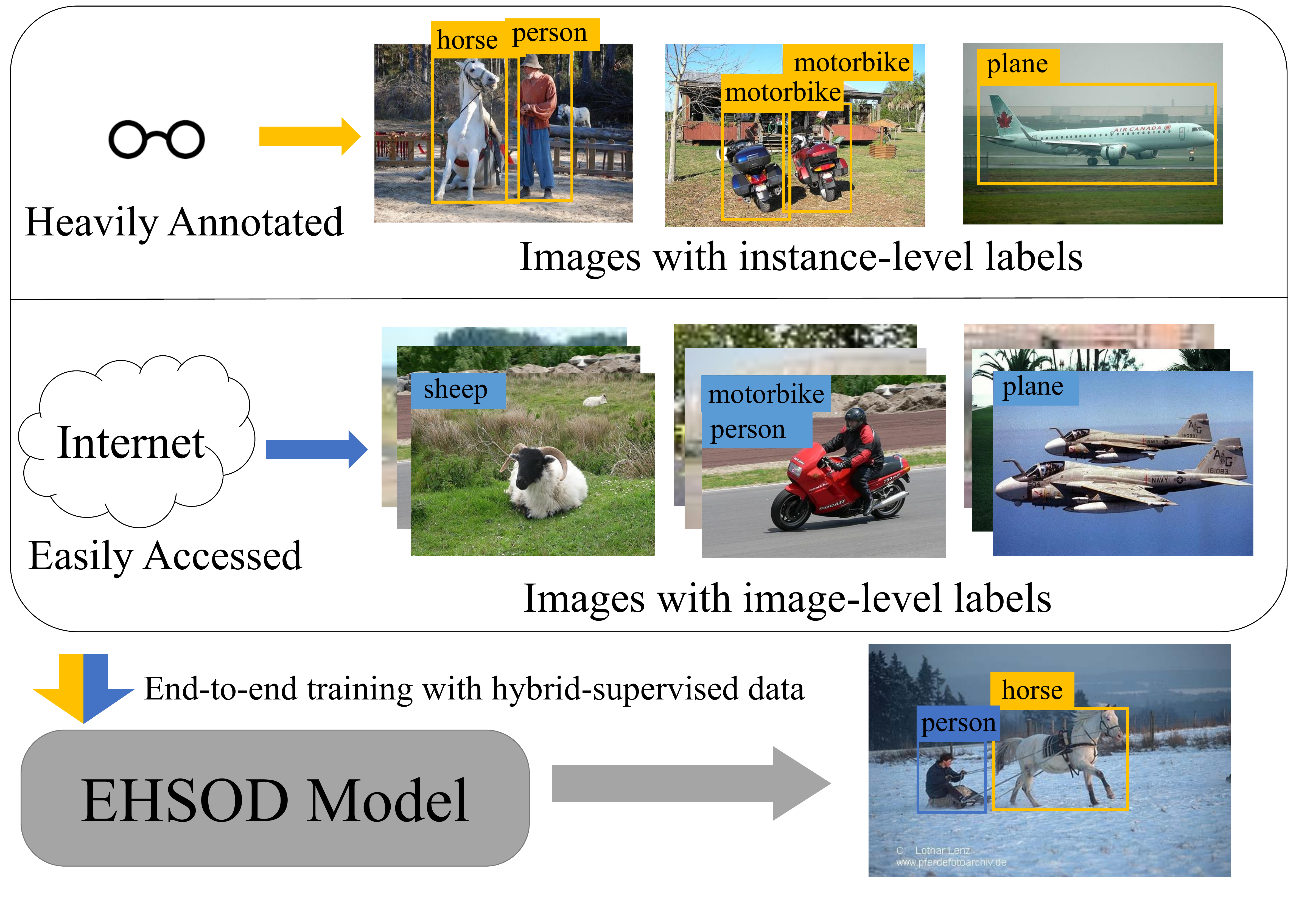}
\par\end{centering}
\caption{\label{fig:intro-graph}An illustration of our hybrid-supervised object
detection task. Hybrid-supervised object detection problem focuses
on training a good detector with a) limited fully-annotated data with
bounding-box labels; b) fully utilizing cheap data with image-level
labels. The conventional weakly-supervised methods usually require
alternating iterative strategy while we aim to design an end-to-end
hybrid-supervised object detection which can fully utilize both kinds
of data.}
\end{figure}

Recent advances of object detectors trained on large-scale datasets
with instance-level annotations have shown promising results which
predict both the class labels and the locations of objects in an image
  \parencite{lin2017feature,redmon2017yolo9000,li2018detnet,wang2019region}.
Those detectors are typically trained under full supervision which
requires huge manual annotations of the objects' locations and categories
for a large number of training images. However, it is expensive and
time-consuming to recruit annotators for labeling the images. This
becomes more severe when the number of categories is large. Thus,
training a customized good object detector with a limited budget becomes
a crucial problem in community. On the other hand, image-level labels
that indicate the presence of an object can be acquired cheaply even
in large amounts as such labels can be collected easily using an Internet
crawler in an image search engine. Unfortunately, training solely
with weakly-supervised methods yields models of subpar performance
that cannot be reliably used in real life scenarios. As a result,
we study the Hybrid Supervised Object Detection (HSOD) problem focusing
on training a good detector with some fully-annotated data with bounding-box
labels while fully utilizing weakly-annotated data with image-level
labels. Note that this task is different from semi-supervised detection
settings which usually focus on training with some existing categories
with annotated labels and infering on some new categories. 

The current state-of-the-art weakly-supervised/few-shot object detection
methods usually require generating pseudo labels and updating the
detector iteratively. Most of the previous methods follow the Multiple
Instance Learning (MIL) pipelines \parencite{cinbis2016weakly,li2016weakly,jie2017deep}:
images are decomposed into object proposals and the learning process
iteratively alternates between re-localizing objects given the current
detector and re-training the detector given the current object locations.
During re-localization, different kinds of scoring systems are adopted
to select the best proposal for an object class in each image. The
recently proposed end-to-end PCL method requires mining instance labels
for online instance classifier refinement  \parencite{tang2017multiple,tang2018pcl}.
Although some promising results have been obtained for Weakly Supervised
Object Detector (WSOD)  \parencite{zhang2018w2f,wan2018min,wei2018ts2c,wan2019c,li2019weakly,kosugi2019object},
they are not comparable to those of fully supervised ones to meet
the standard of deployment on the product.

Moreover, this kind of iterative training paradigm can easily get
stuck in a local minimum, and are therefore unstable, which means
it requires careful hyper-parameter tuning in mining good pseudo labels
for each round of training. The weakly-supervised detector will easily
fail when bad proposals are adding into training. When using a new
dataset, many human efforts are further required to find a new set
of good hyper-parameters. This situation is more severe when the number
of categories and the dataset is large which hinders their usage in
industry. To solve these issues, we seek to find a useful end-to-end
object detection system which can be trained only once on both kinds
of data, which nearly increase no extra hyper-parameters compared
to the fully-supervised counterpart. 

In this work, we present EHSOD for designing an end-to-end hybrid-supervised
object detection network. The proposed method incorporates end-to-end
joint training with both kinds of data and fully utilizes relevant
information of the image-level labeled data to reach better performance.
Specifically, based on a standard two-stage detector  \parencite{ren2015faster},
we proposed two modules to upgrade the existing system and manage
to learn from both kinds of data. Stacked on an ImageNet pretrained
ResNet, a CAM-RPN is proposed to localize the foreground proposals
guided by a class activation heat-map (CAM) \parencite{zhou2016learning}.
The CAM is jointly trained by the image-level labels and bounding-box
annotations within the feature hierarchy of FPN  \parencite{lin2017feature}
and provides further information for Region Proposal Network (RPN)
to select proposals. Furthermore, we design a hybrid-supervised cascade
module to progressively refine the bounding-box position and improve
classification accuracy. This module is a sequence of cascaded hybrid-supervised
heads that contains a regular RCNN head as in  \citeauthor{Lin2017a}
and an auxiliary Multiple Instance Detection (MID) head as in  \citeauthor{bilen2016weakly}.
The hybrid-supervised head is compatible with both kinds of data and
can use image-level data to enhance the learning of classification. 

The training of the resulting detection network follows a standard
procedure of two-stage detection  \parencite{he2018rethinking}. The
final detection system is greatly enhanced by abundant relevant image-level
information and the performance is then boosted by sharing and distilling
essential information across weakly/fully-supervised data.

Extensive experiments are conducted on the widely used detection benchmarks,
including Pascal VOC  \parencite{everingham2015pascal} and MS-COCO
 \parencite{lin2014microsoft}. The proposed method outperforms current
state-of-the-art HSOD and WSOD methods, e.g. PCL  \parencite{tang2018pcl}
and BAOD  \parencite{pardo2019baod}. We observe consistent performance
gains on the base detection network FPN  \parencite{lin2017feature}
training with fully-annotated data. In particular, our method achieves
comparable results with fully-supervised methods on multiple object
detection benchmarks, e.g. 37.5\% mAP on MS-COCO using only 30\% fully-annotated
data and 40.0\% mAP with 50\% MS-COCO fully-annotated data (compared
to 37.2\% with FPN trained with whole data).

To sum up, we make the following contributions: 
\begin{itemize}
\item We are among the first to investigate the hybrid-supervised object
detection problem focusing on training on a limited amount of  fully-annotated images
and a large amount of weakly labeled data.
\item By fully exploiting the potential of both kinds information flow with
different kinds of label, we develop EHSOD, a CAM-guided end-to-end
hybrid-supervised object detection system with cascade refinement
which can be trained in an one shot fashion.
\item Extensive experiments demonstrate the effectiveness of the proposed
method and achieve reliable results on multiple object detection benchmarks
by only 30\% fully-annotated data.
\end{itemize}

\section{Related Work}

\begin{figure*}[t]
\begin{centering}
\includegraphics[scale=0.25]{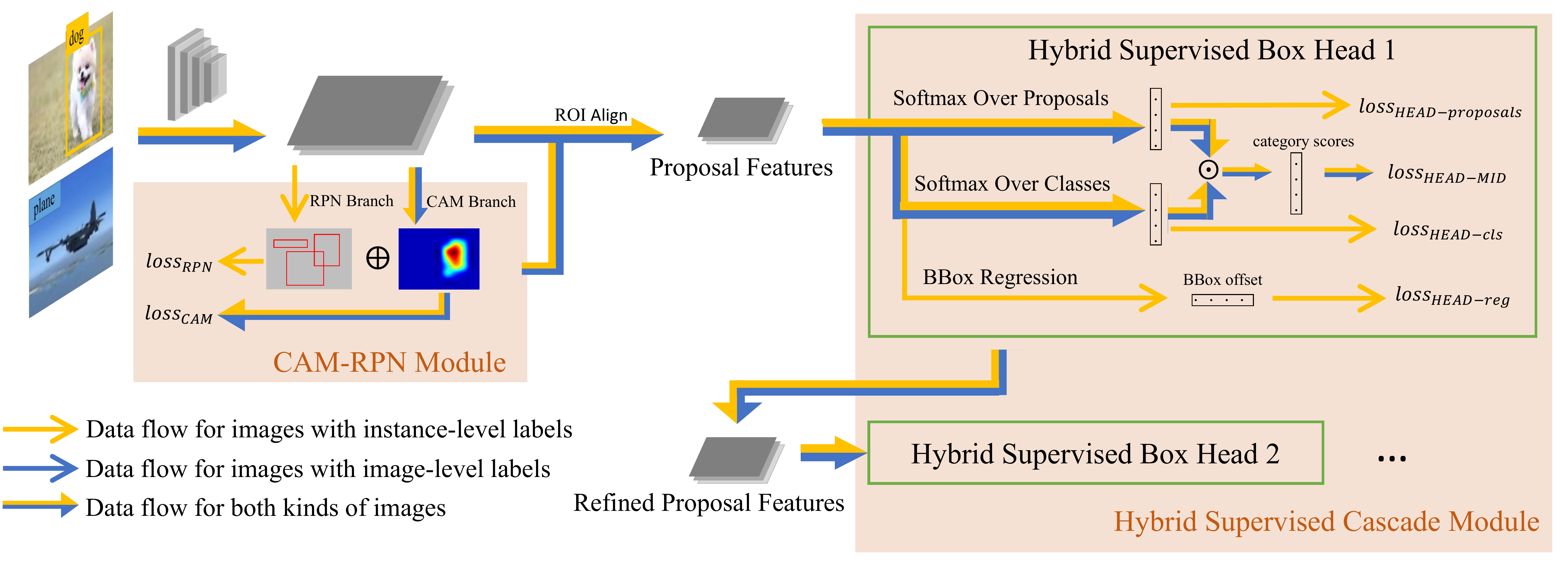}
\par\end{centering}
\caption{\label{fig:framework-graph}An overview of our EHSOD: a hybrid-supervised
object detection framework trained by both image-level labels and
instance-level labels. Stacked on an ImageNet pretrained backbone,
we first proposed a CAM-RPN module to generate the foreground proposals
guided by a class activation heat-map (CAM). The CAM branch is trained
by both kinds of data jointly and provides enhanced objectness score
for RPN to select proposals. Then a sequence of cascaded hybridsupervised
heads further refines the bounding box position and improve classification
accuracy. Within each head, image-level information is effectively
incorporated to enhance the learning of classifiers. Both the bounding-box
labels and the image-level labels are used to improve the performance
of both classification and localization in an end-to-end manner. }
\end{figure*}

\textbf{Fully Supervised Object Detection (FSOD).} Object detection
is a core problem in computer vision. Significant progress has been
made in recent years on FSOD task using CNN. Modern CNN based FSOD
methods may be categorized in two groups: one-stage detection methods
such as SSD and YOLO  \parencite{liu2016ssd,redmon2016you}
and two-stage detection methods such as Faster R-CNN 
and R-FCN  \parencite{dai2016r,ren2015faster,xu2019reasoning,xu2019spatial}. Although these methods have achieved
satisfactoring detection results, the requirement of large-scale bounding-box
annotations may hinder their usage in some budget-aware scenarios.

\textbf{Weakly Supervised Object Detection (WSOD). }WSOD aims at training
a detector with only image-level labels, and received extensive attention
from both academia and industry. Some classical methods formulate
WSOD as a MIL problem  \parencite{cinbis2016weakly,li2016weakly,jie2017deep,zhang2018zigzag},
which treats each training image as a bag of candidate instances and
work in an iterative way to find the positive proposals and train
a detector. In contrast to those interative MIL methods, some works
try to construct end-to-end MIL models for WSOD  \parencite{tang2017multiple,tang2018pcl,bilen2016weakly,diba2017weakly}.
Another kind of methods mine pseudo labels from the location information
obtained by the WSOD approaches to learn a supervised detector  \parencite{zhang2018w2f,zhang2018zigzag,shen2018generative}.
These approaches easily fail without heavy hyper-parameter tuning
and hardly achieve the performance requirement of most real-life applications.

\textbf{Hybrid Supervised Object Detection (HSOD).} HSOD aims at using
a small number of fully-supervised data and a large number of weakly-supervised
data to train a high-performan detector.  \citeauthor{yan2017weakly}
developed an Expectation-Maximization (EM) based method for both WSOD
and HSOD.  \citeauthor{pardo2019baod} used teacher-student learning
method to solve the HSOD problem. These methods work in an multi-step
way, which heavily relys on hyperparameter tuning for mining high-quality
pseudo object labels. We aims to solve this issue by construcing an
end-to-end HSOD model that adds some weakly-supervised branches to
the standard FSOD models without changing its structure. The whole
model can be trained in an one shot fashion jointly on both kinds
of data nearly without extra hyper-parameter tuning. It should be
noted that our work differents from previous semi-supervised detection
works that transfer knowlege from fully-supervised categories to weakly-supervised
categories  \parencite{tang2016large,uijlings2018revisiting}. Our
work focus on the setting where both instance-level labels and image-level
labels are in the same categories.

\section{The Proposed Approach}

\textbf{Overview.} In this paper, we introduce EHSOD: a hybrid-supervised object detection
framework to develop a general detection model to incorporate both
image-level and bounding-box information. An overview of our EHSOD
can be found in Figure \ref{fig:framework-graph}. The proposed EHSOD
is stacked on an ImageNet pretrained backbone to extract feature.
A CAM-RPN is used to propose the foreground proposals guided by a
class activation heat-map (CAM). The CAM is generated by two convolutional
layers trained by the image-level labels and provide further information
for RPN to select proposals. To generate classification score and
location offset for each proposal, a sequential of cascaded hybrid-supervised
heads progressively refines the bounding box position and improves
classification accuracy. Within each hybrid-supervised head, image-level
information is incorporated to enhance the learning of classifiers.
Both the bounding-box labels and the image-level labels are used to
improve the performance of both classification and localization in
an end-to-end manner. 

\subsection{CAM-RPN Module}

Conventional RPN in two-stage detection framework generates proposals
based on predefined anchors. It has one classification layer to outputs
the foreground scores and a regression layer to produce bounding-box
offsets which is effective on fully-annotated data. We further improve
the RPN module to utilize the image-level information.

Inspired from some works on WSOD and segmentation  \parencite{zhou2016learning,diba2017weakly},
class activation heat-map (CAM) presents the activated area in the
feature map of an CNN and can help to locate the object. To enable
end-to-end training and improve the usage of both kinds of data, we
make following modifications on CAM. Let $\mathbf{f}=\{f_{l}\}_{l=1}^{4},f_{l}\in\mathbf{\mathbb{R}}^{W_{l}\times H_{l}\times D}$
be the four different feature maps with different resolutions in the
FPN hierarchy extracted from the output of each stage of the ResNet
backbone network. We use the same design of RPN (3x3 conv and 1x1
conv) to transform the $f_{l}$ to a class activation heat-map $A_{l}$
with $C$ channels ($C$ is the number of the categories). By adopting
global average pooling over $A_{l}$ and then performing the softmax
operation, we obtain a vector $\boldsymbol{y}_{CAM}=[y_{1},...,y_{C}]$
presenting the predicted probability of each category in an image.
Thus, we can construct a multi-label classification loss over the
ground truth of image-level information:
\[
\mathcal{L}_{CAM-cls}=-\sum_{c}\left\{ y_{c}^{*}logy_{c}+(1-y_{c}^{*})log(1-y_{c})\right\} ,
\]
where $\mathcal{L}_{CAM-cls}$ is the binary cross-entropy loss and
$y_{c}^{*}\in\left\{ 0,1\right\} $ is the label of the presence
or absence of class $c$.

Note that the conventional CAM is purely learned from image-level
information. To better reflect the position of objects of the generated
CAM, we fully make use of the ground truth bounding-box. We first
generate a ground truth heat-map for each $A_{l}$ from the annotated
bounding-boxes. Given an object instance of category $c$ and its
bounding-box coordinates $b=(x,y,w,h)$, we map it to the corresponding
feature map scale $s$ for generating the ground truth map of $A_{s}(c)$
(the $cth$ channel of $A_{s}$). The mapped box coordinates is denoted
as $b_{m}=(x^{s},y^{s},w^{s},h^{s})$. For the ground truth map of
$A_{s}(c)$, we define the positive region $b_{m}^{p}=(x^{s},y^{s},\sigma w^{s},\sigma h^{s})$
as the proportional region of $b_{m}$ by a constant scale factor
$\sigma$, the remaining region of $b_{m}$ as the ignoring region
$b_{m}^{i}=b_{m}\backslash b_{m}^{p}$, and the whole map excluding
the $b_{m}$ as the negative region. According to the ground truth
heat-map $A^{*}=\{A_{l}^{*}\}_{l=1}^{4}$, we add a pixel-level segmentation
loss between the generated CAM:{\scriptsize{}
\begin{align*}
 & \mathcal{L}_{CAM-seg}\\
 & =-\alpha\sum_{l}\sum_{c}\sum_{i,j}\{A_{l}^{*}(c,i,j)(1-A_{l}(c,i,j))^{\gamma}logA_{l}(c,i,j)\\
 & +(1-A_{l}^{*}(c,i,j))(A_{l}(c,i,j))^{\gamma}log(1-A_{l}(c,i,j))\},
\end{align*}
}where $\mathcal{L}_{CAM-seg}$ is the pixel-level focal loss, $A_{l}(c,i,j)$
is the predicted probability on pixel $(i,j)$ of $c$th channel of
$A_{l}$obtained by performing an element-wise sigmoid function on
the CAM, and $A_{l}^{*}(c,i,j)$ is the corresponding ground truth
on pixel $(i,j)$. For each proposal, we can calculate a objectness
score from the CAM by performing a softmax operation over channels
on the matched $A_{l}$ and calculate the mean value over the proposal's
region on the obtained single-channel CAM. The summation of the CAM
objectness score and the confidence score from the RPN classification
layer is used to perform non-maximum suppression (NMS) to select the
best proposals for the next stage of hybrid-supervised cascade module.

To fully incorporate both kinds of data, the loss function of CAM-RPN
is formulated as the weighted summation of the following four loss
items: 

\begin{align*}
\mathcal{L_{\mathrm{\mathit{CAM-RPN}}}=} & \text{\ensuremath{\alpha}}_{1}\mathcal{L}_{CAM-cls}+\alpha_{2}\mathcal{L}_{CAM-seg}\\
 & +\alpha_{3}\mathcal{L}_{RPN-cls}+\alpha_{4}\mathcal{L}_{RPN-reg},
\end{align*}

Where the$\mathcal{L}_{RPN-cls}$ and $\mathcal{L}_{RPN-reg}$ are
the regular RPN losses for predicting foreground confidence score
and bounding-box offets as in \cite{Ren2015}. During training, the
$\mathcal{L}_{CAM-cls}$ and $\mathcal{L}_{CAM-seg}$ are calculated
from both kinds of data, while the $\mathcal{L}_{RPN-cls}$ and $\mathcal{L}_{RPN-reg}$
are only gererated by fully-supervised data. The detailed computational
flowchart is shown in Figure 3.

\begin{figure}
\begin{centering}
\includegraphics[scale=0.15]{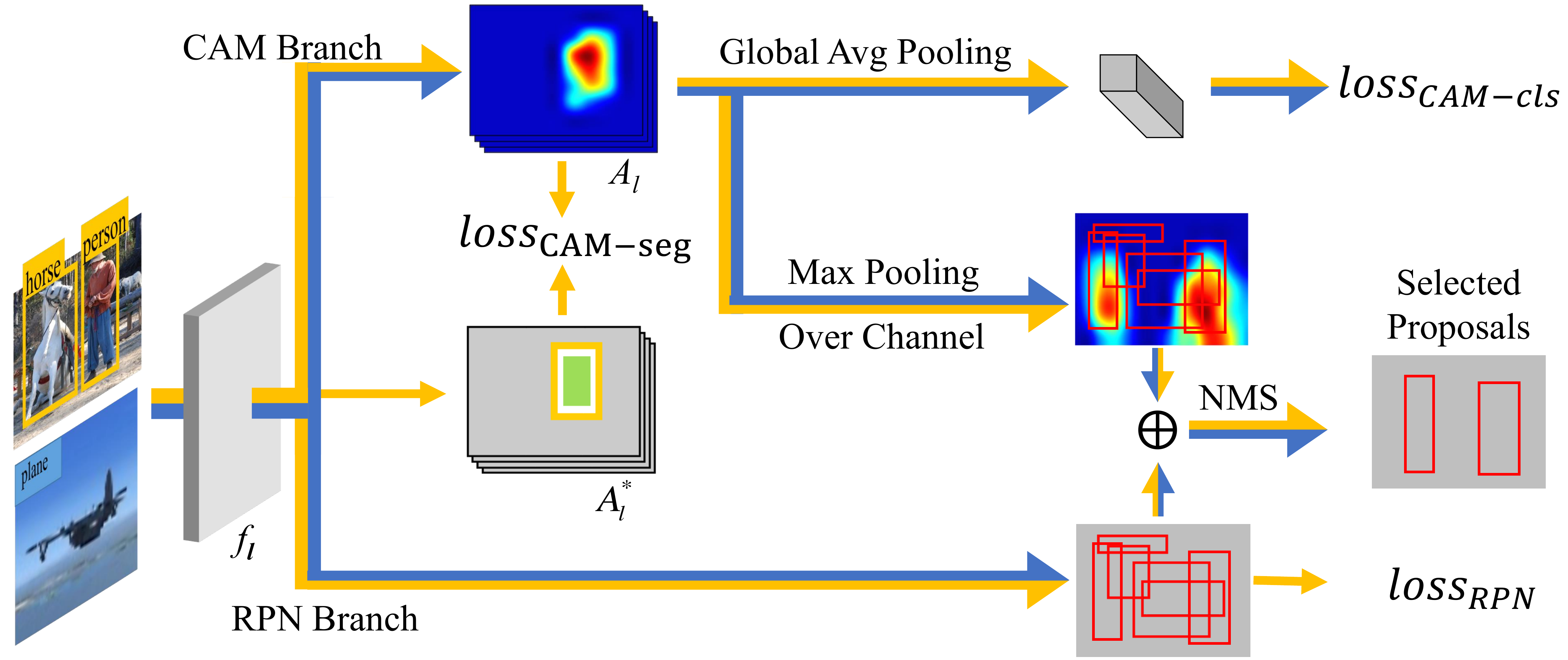}
\par\end{centering}
\caption{Detailed flowchart of the CAM-RPN Module. The whole module includes:
a CAM branch trained by both kinds of data; a RPN branch trained only
by fully-supervised data. The CAM branch is supervised by a image-level
classification loss and a pixel-level bounding-box segmentation loss.
The RPN branch generates a set of proposals and their confidence scores.
Then another enhanced objectness score of each proposal is calculated
by averaging the corresponding region on the CAM. The summation of
these two scores is used as the final foreground score for each proposal,
which is used to select the best proposals for the next stage.}
\end{figure}

\begin{figure*}
	\begin{centering}
		\includegraphics[height=2.45cm]{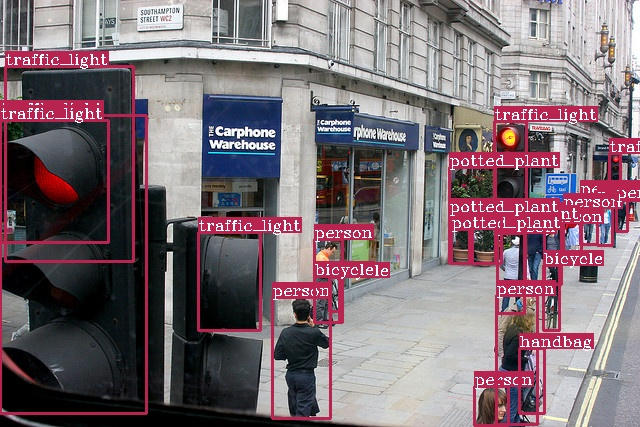}\includegraphics[height=2.45cm]{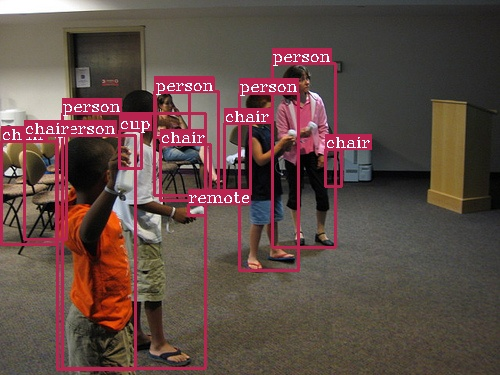}\includegraphics[height=2.45cm]{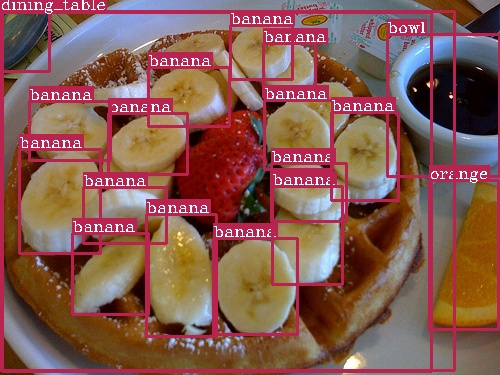}\includegraphics[height=2.45cm]{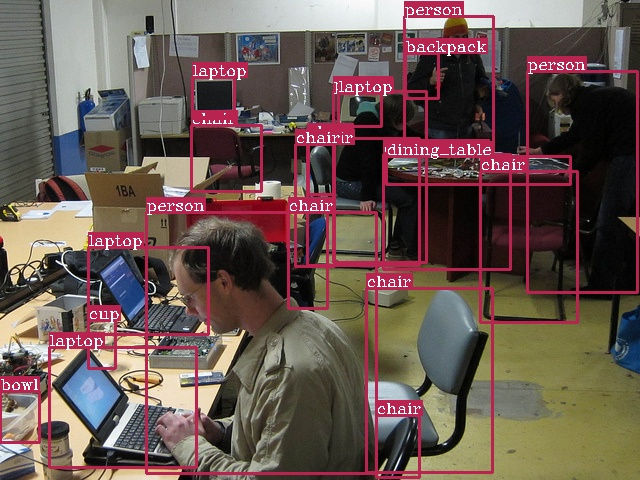}\includegraphics[height=2.45cm]{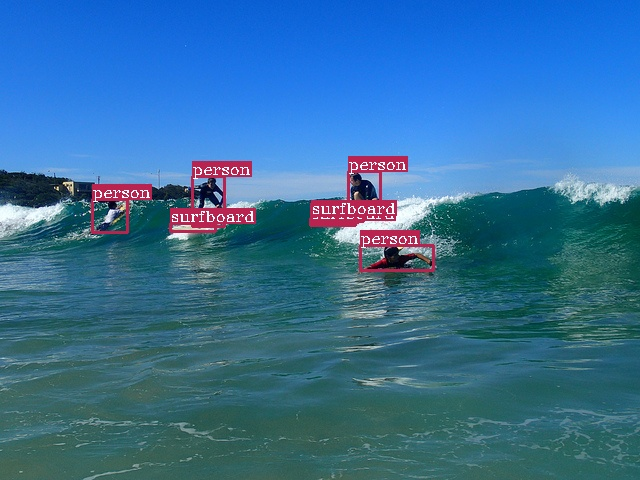}
		\par\end{centering}
	\begin{centering}
		\includegraphics[height=2.45cm]{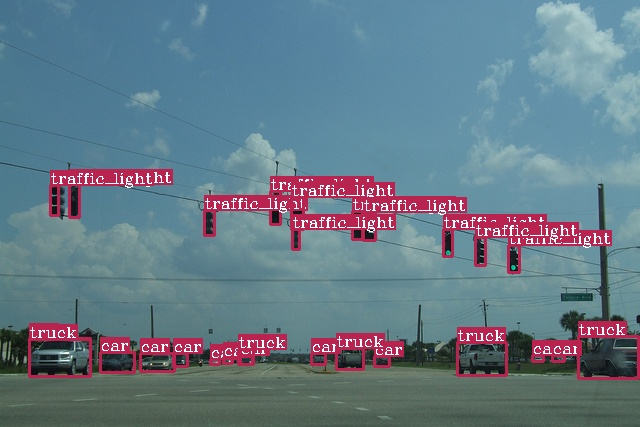}\includegraphics[height=2.45cm]{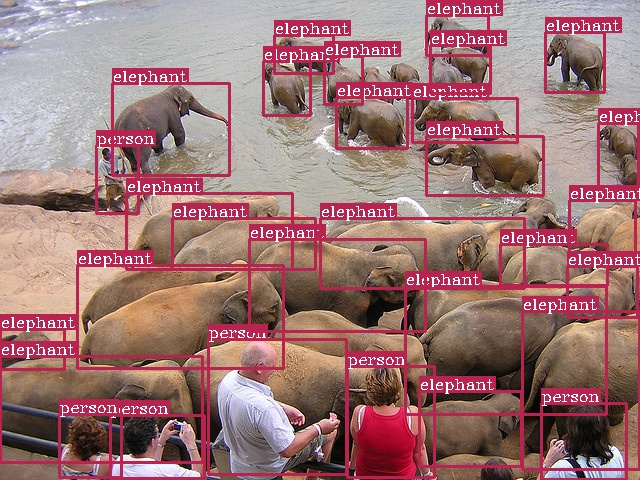}\includegraphics[height=2.45cm]{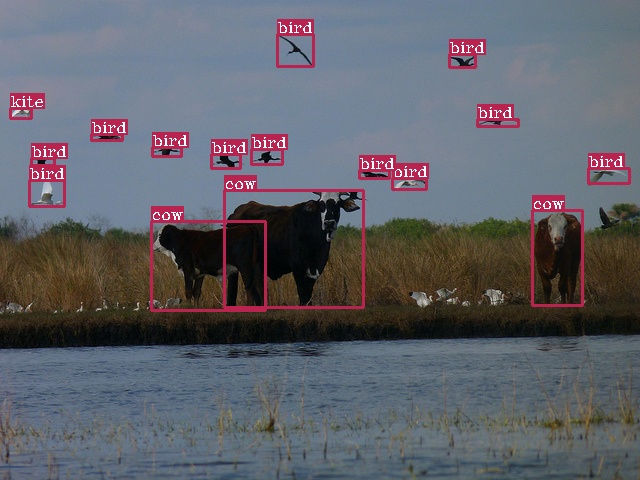}\includegraphics[height=2.45cm]{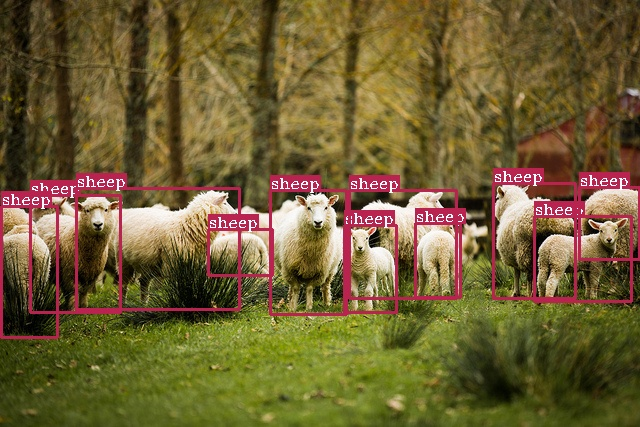}\includegraphics[height=2.45cm]{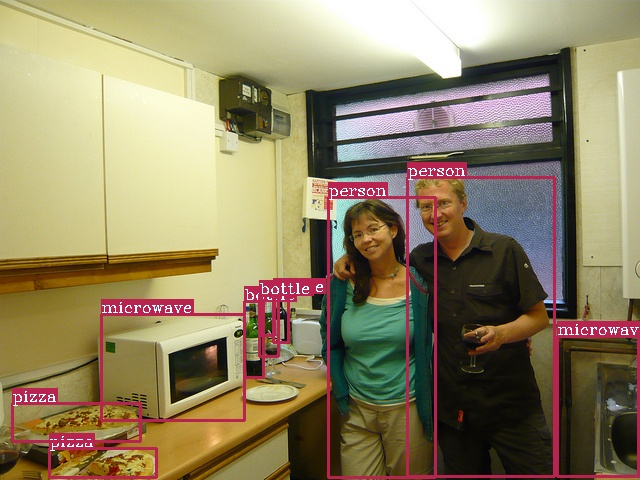}
		\par\end{centering}
	\caption{\label{fig:Qualitative-comparison-of}Qualitative examples of our
		EHSOD trained on 30\% fully-supervised COCO data. EHSOD can detect
		tiny and occusion objects due to the help of image-level information. }
\end{figure*}

\subsection{Hybrid Supervised Cascade Module}

In the conventional two-stage detection, RCNN head performs fine-grained
classification and bounding-box refinement. The main purpose of this
module is to exploit the weakly-labeled data to enhance the performance
of the classifier and refine bounding-boxes.

Given the ROI feature $X=\{\boldsymbol{x}_{1},\boldsymbol{x}_{2},...\boldsymbol{x}_{n}\}$
and $\boldsymbol{x}_{i}\in\mathbb{R}^{D}$ from the $n$ proposals
by the previous CAM-RPN module, like conventional RCNN head for fully-supervised
detector, we use one regression branch to predict the offset of bounding-box
and one classification branch to predict the classification score
$s_{ic}$ for proposal $i$ and categories $c$. Inspired by the network
proposed in WSDNN  \parencite{bilen2016weakly}, we further add another
new proposal-confidence branch to calculate the confidence score for
each proposal. A fully connected layer takes $X$ as input and outputs
confidence score $p_{ic}$ for proposal $i$ on category $c$. Different
from the classification branch, the softmax operation is taken over
the proposals. Thus $\boldsymbol{p}_{:c}$ is a term that ranks all
the proposals with the probability of containing category $c$ while
$\boldsymbol{s}_{i}$ are the probability of proposal i belongs to
each category. Thus, for each image, the probability of containing
an object with category $c$ can be calculated as:
\[
g_{c}=\sum_{i}p_{ic}s_{ic},
\]
which can be also written as an elementary-wise matrix product $\boldsymbol{p}\odot\boldsymbol{s}$
and sum on all the proposals. Softmax is not performed at this step
as images are allowed to contain more than one object class. Note
that the $s_{ic}$ is only calculated by softmax operation over the
logits of foreground object classes at this step. 

From the image-level labels (both from weakly-supervised data and
fully-supervised data), we can train the classification branch and
proposal-confidence branch jointly by a multiple instance detection
(MID) loss:

\[
\mathcal{L}_{HEAD-MID}=-\sum_{c}\left\{ y_{c}logg_{c}+(1-y_{c})log(1-g_{c})\right\} ,
\]

where $\mathcal{L}_{HEAD-MID}$ is the binary cross-entropy loss and
$g_{c}$ is the predicted score of $c$ category. Furthermore, for
each image with bounding-box annotations, we can learn the proposal-confidence
branch with supervision. After assigning category labels to all proposals,
we set the ground truth$p_{ij}^{*}$ as $0$ for proposal $i$ assigned
with background class and set the ground truth $p_{ij}^{*}$ as $1/N_{j}$
for proposal $i$ assigned with category $j$ ($N_{j}$ is the total
number of proposals assigned with category $j$). The resulting $\mathbf{\mathit{\boldsymbol{p}}}_{:j}^{*}$
is set as the ground truth of $\boldsymbol{p}_{:j}$. Thus, we can
train the proposal-confidence branch by a cross-entropy loss:

\[
\mathcal{L_{\mathrm{\mathit{HEAD-\mathit{proposals}}}}=}-\frac{1}{R}\sum_{j}\sum_{i}p_{ij}^{*}logp_{ij},
\]

Where $R$ is total number of region proposals. To fully utilize both
kinds of data, the loss function of a hybrid-supervised head is formulated
as the weighted summation of the following four loss items:

\begin{align*}
\mathcal{L_{\mathrm{\mathit{HS-head}}}=} & \beta_{1}\mathcal{L}_{HEAD-MID}+\beta_{2}\mathcal{L}_{HEAD-proposals}\\
 & +\beta_{3}\mathcal{L}_{HEAD-cls}+\beta_{4}\mathcal{L}_{HEAD-reg},
\end{align*}

Where the $\mathcal{L}_{HEAD-cls}$ and $\mathcal{L}_{HEAD-reg}$
are the regular bounding-box losses as in \cite{Ren2015}. During
traing, the $\mathcal{L}_{HEAD-MID}$ is calculated from both kinds
of data, while the $\mathcal{L}_{HEAD-proposals}$, $\mathcal{L}_{HEAD-cls}$,
and $\mathcal{L}_{HEAD-reg}$ are only gererated by fully-supervised
data. The detail computational flowchart is shown in Figure 2.

Note that the performance of the bounding-box regression branch is
very weak since we have little data to train it. To further refine
the bounding-box position, we adopt the cascade-structure with increasing
IoU threshold. Empirically, we found that a sequence of three cascade
works very well and can boost the localization performance of the
detector.

\subsection{Training EHSOD}

Having discussed the EHSOD architecture in the previous section, here
we explain how the model is trained. The prposed EHSOD framework is
optimized in an end-to-end fashion using a multi-task loss. Apart
from the conventional loss of cascade detection network \cite{Cai2018},
we also introduce two losses $\mathcal{L}_{CAM-cls}$ and $\mathcal{L}_{CAM-seg}$
for CAM learning and two losses $\mathcal{L}_{HEAD-MID}$ and $\mathcal{L}_{HEAD-proposals}$
for hybrid-supervised head learning. They are jointly optimized by
the following weighted summation of all losses:

\[
\mathcal{L}=\lambda_{0}\mathcal{L}_{CAM-rpn}+\sum_{i}\lambda_{i}\mathcal{L_{\mathrm{\mathit{HS-head-i}}}}.
\]

Where $\mathcal{L_{\mathrm{\mathit{HS-head-i}}}}$is the loss for
the $ith$ head. In practice, we found that the convergence of the
model is very fast and we can train the model with the default setting
of a two-stage detection network such as FPN  \parencite{Lin2017a}.

\section{Experiments}

\begin{table}
\begin{centering}
\tabcolsep 0.01in{\footnotesize{}}%
\begin{tabular}{c|ccc}
\hline 
{\footnotesize{}Method} & {\footnotesize{}Backbone} & {\footnotesize{}AP} & {\footnotesize{}AP$_{50}$}\tabularnewline
\hline 
{\footnotesize{}SSD  \parencite{Liu2016}} & {\footnotesize{}VGG16} & {\footnotesize{}26.8} & {\footnotesize{}46.5}\tabularnewline
{\footnotesize{}RetinaNet  \parencite{lin2017focal}} & {\footnotesize{}Res50} & {\footnotesize{}35.6} & {\footnotesize{}54.7}\tabularnewline
{\footnotesize{}Faster R-CNN  \parencite{Ren2015}} & {\footnotesize{}Res50} & {\footnotesize{}32.6} & {\footnotesize{}53.1}\tabularnewline
{\footnotesize{}FPN  \parencite{Lin2017a}} & {\footnotesize{}Res50 } & {\footnotesize{}35.9} & {\footnotesize{}56.9}\tabularnewline
{\footnotesize{}FSAF  \parencite{zhu2019feature}} & {\footnotesize{}Res50 } & {\footnotesize{}37.2} & {\footnotesize{}57.2}\tabularnewline
{\footnotesize{}AlignDet  \parencite{chen2019revisiting}} & {\footnotesize{}Res50 } & \textbf{\footnotesize{}37.9} & \textbf{\footnotesize{}57.7}\tabularnewline
\hline 
{\footnotesize{}EHSOD w }\textbf{\footnotesize{}30\%}{\footnotesize{}
Fully supervised img} & {\footnotesize{}Res50 } & {\footnotesize{}35.3} & {\footnotesize{}54.2}\tabularnewline
{\footnotesize{}EHSOD w }\textbf{\footnotesize{}50\%}{\footnotesize{}
Fully supervised img} & {\footnotesize{}Res50 } & {\footnotesize{}37.8} & {\footnotesize{}56.5}\tabularnewline
\hline 
{\footnotesize{}SSD  \parencite{Liu2016}} & {\footnotesize{}Res101} & {\footnotesize{}31.2} & {\footnotesize{}50.4}\tabularnewline
{\footnotesize{}RetinaNet  \parencite{lin2017focal}} & {\footnotesize{}Res101} & {\footnotesize{}37.7} & {\footnotesize{}57.2}\tabularnewline
{\footnotesize{}Faster R-CNN  \parencite{Ren2015}} & {\footnotesize{}Res101} & {\footnotesize{}34.9} & {\footnotesize{}55.7}\tabularnewline
{\footnotesize{}FPN  \parencite{Lin2017a}} & {\footnotesize{}Res101} & {\footnotesize{}37.2} & {\footnotesize{}59.1}\tabularnewline
{\footnotesize{}FSAF  \parencite{zhu2019feature}} & {\footnotesize{}Res101} & {\footnotesize{}39.3} & {\footnotesize{}59.2}\tabularnewline
{\footnotesize{}AlignDet  \parencite{chen2019revisiting}} & {\footnotesize{}Res101} & \textbf{\footnotesize{}39.8} & \textbf{\footnotesize{}60.0}\tabularnewline
\hline 
{\footnotesize{}EHSOD w }\textbf{\footnotesize{}30\%}{\footnotesize{}
Fully supervised img} & {\footnotesize{}Res101} & {\footnotesize{}37.5} & {\footnotesize{}56.8}\tabularnewline
{\footnotesize{}EHSOD w }\textbf{\footnotesize{}50\%}{\footnotesize{}
Fully supervised img} & {\footnotesize{}Res101} & \textbf{\footnotesize{}40.0} & {\footnotesize{}59.4}\tabularnewline
\hline 
\end{tabular}{\footnotesize\par}
\par\end{centering}
\caption{\label{tab:Results on coco}Comparison with Fully Supervised Object
Detection methods on MS-COCO. The competing methods are trained with
full data (1x schedule and no multi-scale training/testing). EHSOD
trained with only 30\%/50\% fully-supervised images can reach comparable
performance with fully-supervised object detection methods.}
\end{table}

\textbf{Datasets and Evaluations. }We evaluate the performance of
our proposed EHSOD method on two common detection benchmarks: the
PASCAL VOC 2007  \parencite{everingham2015pascal}, and the MS-COCO
2017 dataset  \parencite{lin2014microsoft}. The PASCAL VOC 2007 has
9,962 images with 20 categories. For PASCAL VOC 2007 , we choose trainval
set (5,011 images) for training and choose the test set (4,952 images)
for testing. The MS-COCO dataset has 80 object classes, which is divided
into train set (118K images), val set (5K images) and test set (20K
unannotated images). We train our model on the MS-COCO train set and
test our model on the val set. 

For the hybrid-supervised setting, we select a proportion of training
images randomly as the fully-supervised training data, the remaining
training images are used as weakly-supervised training data. We employ
the standard mean Average Precision (mAP) metric with IoU=0.5 to evaluate
our method on the PASCAL VOC dataset and employ mAP@{[}.5, .95{]}
on the MS-COCO dataset. 

\textbf{Implementation Details. }We use the popular FPN  \parencite{Lin2017a}
as our baseline detector and implement the EHSOD network based on
it. ImageNet pretained backbone is used as the backbone network. We
use a sequence of three cascaded heads with increasing IoU threshold
in the hybrid-supervised cascade module. Thus, our EHSOD network have
four stages in total, one CAM-RPN for generating proposals and three
heads for detection with IoU threshold $\left\{ 0.5,0.6,0.7\right\} $.
We set the loss weights $\alpha_{1}$and $\alpha_{2}$ in $\mathcal{L}_{CAM-RPN}$
to 0.1 and 0.2 respectively, set the loss weights $\lambda_{1}$,
$\lambda_{2}$ and $\lambda_{3}$ for three hybrid-supervised heads
to $1$, $0.5$. $0.25$ respectively, and set all the other loss
weights to $1$. The scale factor $\sigma$ for generating the positive
region of the ground truth CAM is set to 0.8. The hyper-parameters
$\alpha$ and $\gamma$ for focal loss in the $\mathcal{L}_{CAM-seg}$
are set to 0.25 and 2 respectively. No data augmentation was used
except standard horizontal image flipping. 

During both training and testing, we resize the input image such that
the shorter side has 600 pixels and 800 pixels for the PASCAL VOC
dataset and the MS-COCO dataset respectively. All experiments are
conducted on a single server with 8 Tesla V100 GPUs by using the Pytorch
framework. For training, SGD with weight decay of 0.0001 and momentum
of 0.9 is adopted to optimize all models. For the PASCAL VOC dataset,
the batch size is set to be 8 with 4 images on each GPU, the initial
learning rate is 0.005, reduce by 0.1 at epoch 9 during the training
process. For the MS-COCO dataset, the batch size is set to be 16 with
2 images on each GPU, the initial learning rate is 0.01, reduce by
0.1 at epoch 8 and 11 during the training process. We only train \textbf{12
epochs for all models} in an end-to-end manner. \textbf{Multi-scale
training/testing is not used} for all the models. 

\begin{figure}
\begin{centering}
{\scriptsize{}\includegraphics[scale=0.4]{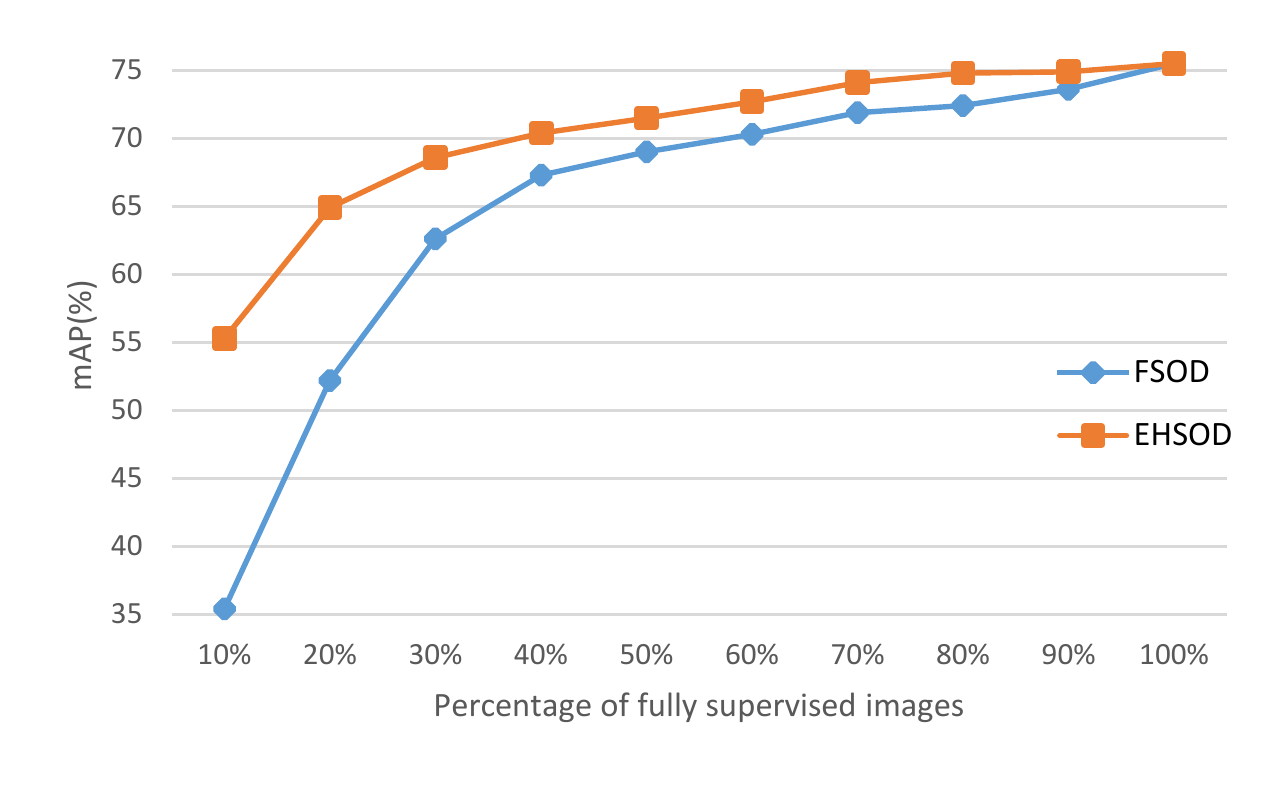}}{\scriptsize\par}
\par\end{centering}
\caption{\label{fig:Comparison betwwen FSOD and HSOD}Comparison between our
EHSOD (Orange) and its fully-supervised version (FSOD: Blue) on different
portions of data of PASCAL VOC07. The blue line is its fully-supervised
object detection counterpart. Note that the image-level information
of the fully-supervised data is still used in the blue line model.
The backbone is ResNet-50.}
\end{figure}

\begin{table*}
\begin{centering}
\tabcolsep 0.02in{\scriptsize{}}%
\begin{tabular}{l|l|cccccccccccccccccccc|c}
\hline 
{\scriptsize{}data} & {\scriptsize{}Method} & {\scriptsize{}aero} & {\scriptsize{}bike} & {\scriptsize{}bird} & {\scriptsize{}boat} & {\scriptsize{}bottle} & {\scriptsize{}bus} & {\scriptsize{}car} & {\scriptsize{}cat} & {\scriptsize{}chair} & {\scriptsize{}cow} & {\scriptsize{}table} & {\scriptsize{}dog} & {\scriptsize{}horse} & {\scriptsize{}mbike} & {\scriptsize{}person} & {\scriptsize{}plant} & {\scriptsize{}sheep} & {\scriptsize{}sofa} & {\scriptsize{}train} & {\scriptsize{}tv} & {\scriptsize{}mAP}\tabularnewline
\hline 
\multirow{2}{*}{{\scriptsize{}10\%}} & {\scriptsize{}BAOD  \parencite{pardo2019baod} } & {\scriptsize{}51.6} & {\scriptsize{}50.7} & {\scriptsize{}52.6} & {\scriptsize{}41.7} & {\scriptsize{}36.0} & {\scriptsize{}52.9} & {\scriptsize{}63.7} & {\scriptsize{}69.7} & {\scriptsize{}34.4} & {\scriptsize{}65.4} & {\scriptsize{}22.1} & {\scriptsize{}66.1} & {\scriptsize{}63.9} & {\scriptsize{}53.5} & {\scriptsize{}59.8} & {\scriptsize{}24.5} & {\scriptsize{}60.2} & {\scriptsize{}43.3} & {\scriptsize{}59.7} & {\scriptsize{}46.0} & {\scriptsize{}50.9}\tabularnewline
 & {\scriptsize{}Our EHSOD} & {\scriptsize{}60.6} & {\scriptsize{}65.2} & {\scriptsize{}55.0} & {\scriptsize{}35.4} & {\scriptsize{}32.8} & {\scriptsize{}66.1} & {\scriptsize{}71.3} & {\scriptsize{}75.3} & {\scriptsize{}38.4} & {\scriptsize{}54.1} & {\scriptsize{}26.5} & {\scriptsize{}71.7} & {\scriptsize{}65.0} & {\scriptsize{}67.8} & {\scriptsize{}63.0} & {\scriptsize{}27.7} & {\scriptsize{}52.6} & {\scriptsize{}48.6} & {\scriptsize{}70.9} & {\scriptsize{}57.3} & \textbf{\scriptsize{}55.3}\tabularnewline
\hline 
\multirow{2}{*}{{\scriptsize{}20\%}} & {\scriptsize{}BAOD  \parencite{pardo2019baod} } & {\scriptsize{}57.0} & {\scriptsize{}62.2} & {\scriptsize{}60.0} & {\scriptsize{}46.6} & {\scriptsize{}46.7} & {\scriptsize{}60.0} & {\scriptsize{}70.8} & {\scriptsize{}74.4} & {\scriptsize{}40.5} & {\scriptsize{}71.9} & {\scriptsize{}30.2} & {\scriptsize{}72.7} & {\scriptsize{}73.8} & {\scriptsize{}64.7} & {\scriptsize{}69.8} & {\scriptsize{}37.2} & {\scriptsize{}62.9} & {\scriptsize{}48.4} & {\scriptsize{}64.1} & {\scriptsize{}59.1} & {\scriptsize{}58.6}\tabularnewline
 & {\scriptsize{}Our EHSOD} & {\scriptsize{}65.5} & {\scriptsize{}72.3} & {\scriptsize{}66.7} & {\scriptsize{}45.6} & {\scriptsize{}50.8} & {\scriptsize{}72.2} & {\scriptsize{}77.8} & {\scriptsize{}82.2} & {\scriptsize{}44.3} & {\scriptsize{}73.1} & {\scriptsize{}44.8} & {\scriptsize{}79.3} & {\scriptsize{}76.0} & {\scriptsize{}73.0} & {\scriptsize{}73.8} & {\scriptsize{}35.5} & {\scriptsize{}63.0} & {\scriptsize{}62.1} & {\scriptsize{}74.0} & {\scriptsize{}65.5} & \textbf{\scriptsize{}64.9}\tabularnewline
\hline 
\multirow{2}{*}{{\scriptsize{}40\%}} & {\scriptsize{}BAOD  \parencite{pardo2019baod} } & {\scriptsize{}68.6} & {\scriptsize{}71.3} & {\scriptsize{}66.6} & {\scriptsize{}52.5} & {\scriptsize{}53.1} & {\scriptsize{}69.6} & {\scriptsize{}77.7} & {\scriptsize{}77.2} & {\scriptsize{}45.7} & {\scriptsize{}72.7} & {\scriptsize{}54.0} & {\scriptsize{}74.4} & {\scriptsize{}74.6} & {\scriptsize{}74.7} & {\scriptsize{}74.4} & {\scriptsize{}42.4} & {\scriptsize{}66.2} & {\scriptsize{}56.8} & {\scriptsize{}71.7} & {\scriptsize{}65.4} & {\scriptsize{}65.5}\tabularnewline
 & {\scriptsize{}Our EHSOD} & {\scriptsize{}75.8} & {\scriptsize{}78.4} & {\scriptsize{}72.9} & {\scriptsize{}56.7} & {\scriptsize{}55.2} & {\scriptsize{}76.1} & {\scriptsize{}81.3} & {\scriptsize{}83.9} & {\scriptsize{}51.2} & {\scriptsize{}76.2} & {\scriptsize{}60.0} & {\scriptsize{}83.3} & {\scriptsize{}81.5} & {\scriptsize{}77.9} & {\scriptsize{}79.3} & {\scriptsize{}41.2} & {\scriptsize{}68.0} & {\scriptsize{}64.4} & {\scriptsize{}75.2} & {\scriptsize{}68.8} & \textbf{\scriptsize{}70.4}\tabularnewline
\hline 
\multirow{2}{*}{{\scriptsize{}50\%}} & {\scriptsize{}BAOD  \parencite{pardo2019baod} } & {\scriptsize{}70.1} & {\scriptsize{}73.1} & {\scriptsize{}70.4} & {\scriptsize{}52.0} & {\scriptsize{}57.0} & {\scriptsize{}73.1} & {\scriptsize{}79.4} & {\scriptsize{}77.1} & {\scriptsize{}47.4} & {\scriptsize{}77.5} & {\scriptsize{}54.0} & {\scriptsize{}76.6} & {\scriptsize{}73.5} & {\scriptsize{}74.6} & {\scriptsize{}77.1} & {\scriptsize{}43.8} & {\scriptsize{}68.5} & {\scriptsize{}61.3} & {\scriptsize{}73.7} & {\scriptsize{}69.1} & {\scriptsize{}67.5}\tabularnewline
 & {\scriptsize{}Our EHSOD} & {\scriptsize{}73.4} & {\scriptsize{}77.8} & {\scriptsize{}72.9} & {\scriptsize{}57.5} & {\scriptsize{}57.2} & {\scriptsize{}79.5} & {\scriptsize{}81.6} & {\scriptsize{}83.5} & {\scriptsize{}53.7} & {\scriptsize{}79.0} & {\scriptsize{}60.4} & {\scriptsize{}83.5} & {\scriptsize{}81.9} & {\scriptsize{}76.6} & {\scriptsize{}79.7} & {\scriptsize{}45.4} & {\scriptsize{}69.1} & {\scriptsize{}67.3} & {\scriptsize{}77.9} & {\scriptsize{}71.9} & \textbf{\scriptsize{}71.5}\tabularnewline
\hline 
\multirow{2}{*}{{\scriptsize{}60\%}} & {\scriptsize{}BAOD  \parencite{pardo2019baod} } & {\scriptsize{}73.5} & {\scriptsize{}75.3} & {\scriptsize{}72.4} & {\scriptsize{}52.5} & {\scriptsize{}53.3} & {\scriptsize{}76.5} & {\scriptsize{}81.1} & {\scriptsize{}81.0} & {\scriptsize{}51.0} & {\scriptsize{}76.7} & {\scriptsize{}57.9} & {\scriptsize{}76.8} & {\scriptsize{}79.2} & {\scriptsize{}77.0} & {\scriptsize{}79.0} & {\scriptsize{}45.4} & {\scriptsize{}69.3} & {\scriptsize{}63.0} & {\scriptsize{}75.3} & {\scriptsize{}67.2} & {\scriptsize{}69.2}\tabularnewline
 & {\scriptsize{}Our EHSOD} & {\scriptsize{}74.4} & {\scriptsize{}81.3} & {\scriptsize{}72.7} & {\scriptsize{}58.1} & {\scriptsize{}58.9} & {\scriptsize{}82.3} & {\scriptsize{}83.9} & {\scriptsize{}82.9} & {\scriptsize{}54.2} & {\scriptsize{}77.6} & {\scriptsize{}63.5} & {\scriptsize{}82.6} & {\scriptsize{}82.1} & {\scriptsize{}79.6} & {\scriptsize{}80.5} & {\scriptsize{}46.8} & {\scriptsize{}71.2} & {\scriptsize{}71.8} & {\scriptsize{}80.0} & {\scriptsize{}69.9} & \textbf{\scriptsize{}72.7}\tabularnewline
\hline 
\multirow{2}{*}{{\scriptsize{}80\%}} & {\scriptsize{}BAOD  \parencite{pardo2019baod} } & {\scriptsize{}76.7} & {\scriptsize{}76.4} & {\scriptsize{}74.0} & {\scriptsize{}56.8} & {\scriptsize{}62.0} & {\scriptsize{}81.4} & {\scriptsize{}82.1} & {\scriptsize{}84.8} & {\scriptsize{}57.3} & {\scriptsize{}78.2} & {\scriptsize{}61.2} & {\scriptsize{}81.9} & {\scriptsize{}79.3} & {\scriptsize{}78.1} & {\scriptsize{}80.6} & {\scriptsize{}46.8} & {\scriptsize{}73.0} & {\scriptsize{}67.6} & {\scriptsize{}76.9} & {\scriptsize{}71.7} & {\scriptsize{}72.3}\tabularnewline
 & {\scriptsize{}Our EHSOD} & {\scriptsize{}83.1} & {\scriptsize{}82.9} & {\scriptsize{}77.0} & {\scriptsize{}60.6} & {\scriptsize{}63.4} & {\scriptsize{}81.5} & {\scriptsize{}85.2} & {\scriptsize{}86.1} & {\scriptsize{}56.2} & {\scriptsize{}80.5} & {\scriptsize{}65.9} & {\scriptsize{}84.2} & {\scriptsize{}83.1} & {\scriptsize{}79.1} & {\scriptsize{}82.5} & {\scriptsize{}47.8} & {\scriptsize{}73.8} & {\scriptsize{}71.7} & {\scriptsize{}79.7} & {\scriptsize{}72.4} & \textbf{\scriptsize{}74.8}\tabularnewline
\hline 
{\scriptsize{}100\%} & {\scriptsize{}Our EHSOD} & {\scriptsize{}82.5} & {\scriptsize{}82.7} & {\scriptsize{}75.4} & {\scriptsize{}63.3} & {\scriptsize{}63.2} & {\scriptsize{}82.1} & {\scriptsize{}85.8} & {\scriptsize{}86.3} & {\scriptsize{}57.6} & {\scriptsize{}79.5} & {\scriptsize{}67.5} & {\scriptsize{}84.1} & {\scriptsize{}82.6} & {\scriptsize{}80.2} & {\scriptsize{}82.8} & {\scriptsize{}51.5} & {\scriptsize{}73.6} & {\scriptsize{}73.5} & {\scriptsize{}82.7} & {\scriptsize{}74.3} & {\scriptsize{}75.5}\tabularnewline
\hline 
\end{tabular}{\scriptsize\par}
\par\end{centering}
\caption{\label{tab:Comparison betwwen state-of-the-art HSOD}Comparison of
Hybrid Supervised Object Detection methods on VOC07. Both methods
are trained with same settings of hybrid-supervised data. Our method
is trained once with 12 epochs while BAOD has several rounds of training
(each with 10 epochs). It can be found that our method consistently
outperforms the BAOD especially under lower proportions of fully-supervised
data.}
\end{table*}

\textbf{Comparison with Fully Supervised Object Detection methods.
}To show the effectiveness our method in using low-cost annotating
(e.g., weakly-supervised) data to boost the detection performance,
we compare the overall performance of our EHSOD method with its fully-supervised
object detection counterpart. Specifically, we train the same proposed
model without the data-flow of the weakly-supervised data. Note that
the image-level information of the fully-supervised data will still
be used in the model to train the CAM module and the classification
branch/proposal-confidence branch in Hybrid Supervised Cascade Module.
The backbone is ResNet-50. Figure \ref{fig:Comparison betwwen FSOD and HSOD}
shows comparison between EHSOD (Red) and its fully-supervised version
(Blue) on different portions of data. It can be found that our model
can boost the performance mostly in 10\% fully-supervised data. On
Pascal VOC07, EHSOD can significantly increase the performance of
mAP from 35\% to 55\% under only 10\% of fully-supervised data.

In Table \ref{tab:Results on coco}, we further compared our method
with fully object detection training with 100\% data on MS-COCO. The
competing methods are trained with full data. The reported results
use 1x schedule  \parencite{he2018rethinking} and no multi-scale training/testing,
which is under the same setting with us. It can be found that our
method trained with 30\% data has comparable performance with fully-trained
detector such as Faster-RCNN \parencite{ren2015faster}, FPN  \parencite{lin2017feature},
RetinaNet  \parencite{lin2017focal} and SSD  \parencite{liu2016ssd}.
Note that our model with backbone Resnet-101 can reached mAP of 40\%
with only 50\% fully-supervised data. Figure \ref{fig:Qualitative-comparison-of}
further shows some quantitative results for our EHSOD trained with
30\% fully-supervised data on COCO. Our method has a very high accuracy
and can detect very small items such as birds and traffic lights.

\textbf{Comparison with Hybrid Supervised Object Detection method.
}Despite of the limited research works on hybrid-supervised detection
network, we can compare the performance with BAOD  \parencite{pardo2019baod}
with same setting of experiments. BAOD considered an iterative training
scheme by an optimal image/annotation selection and retraining the
detector. Table \ref{tab:Comparison betwwen state-of-the-art HSOD}
compares the mAP$_{50}$ under different setting of fully supervised
data proportion from 10\% to 100\% on Pascal VOC07. Both methods are
under same setting of experiments. Our method is trained once with
12 epochs while BAOD has several rounds of training. 

From Table \ref{tab:Comparison betwwen state-of-the-art HSOD}, it
can be found that our method consistently outperforms the competitor
BAOD. Note that our method is significantly better than BAOD by around
5\% of mAP under 10\%-40\% settings. This demonstrates the effectiveness
of our method in utilizing weakly-supervised data. By observing the
performance gain compared to the baseline method, it can be found
that our method can successfully detect difficult categories such
as aero, bike, bird, bottle and plant. These categories usually suffer
from the problems of tiny-size and occlusion. Our method can alleviate
these problems by the refinement of the hybrid-supervised heads.

\begin{table}
\begin{centering}
\tabcolsep 0.04in{\small{}}%
\begin{tabular}{c|c|c|c}
\hline 
\textit{\small{}VOC 2007} & {\small{}AP$_{50}$} & \textit{\small{}MS-COCO} & {\small{}AP$_{50}$}\tabularnewline
\hline 
{\small{}PCL} & {\small{}48.8} & {\small{}PCL} & {\small{}19.6}\tabularnewline
{\small{}Our EHSOD} & \textbf{\small{}55.3$^{+6.5}$} & {\small{}Our EHSOD} & \textbf{\small{}46.8$^{+27.2}$}\tabularnewline
\hline 
\end{tabular}{\small\par}
\par\end{centering}
\caption{\label{tab:Comparison betwwen WSOD}Comparison of the weakly-supervised
method PCL and our method on VOC07 and MS-COCO. The EHSOD method is
trained with 10\% of fully-supervised data. In MS-COCO, our method
outperforms PCL method by a large margin of 27.2\% in terms of AP$_{50}$.}
\end{table}

\textbf{Comparison with Weakly Supervised Object Detection methods.
}We further compare our approach with weakly-supervised object detection
methods. Current WSOD methods mainly focus on easy detection datasets
such as Pascal VOC, and only PCL  \parencite{tang2018pcl} is tested
on a much harder dataset: MSCOCO. Table \ref{tab:Comparison betwwen WSOD}
shows the comparison between PCL and our method on VOC07 and MS-COCO.
The EHSOD method is trained with 10\% of supervised data. Although
PCL performs well in Pascal VOC, its mAP$_{50}$ in MS-COCO is only
19.4. Our method significantly outperforms PCL by 27.2\%, which implies that our method is superior in harder tasks.

\begin{table}
\begin{centering}
\tabcolsep 0.01in{\scriptsize{}}%
\begin{tabular}{c|c|c|c|c|c}
\hline 
{\scriptsize{}Training with 10\%} & {\scriptsize{}Cascade } & {\scriptsize{}CAM } & {\scriptsize{}Head adds} & {\scriptsize{}Head adds} & \multirow{2}{*}{{\scriptsize{}mAP$_{50}$}}\tabularnewline
{\scriptsize{}fully-supervised data} & {\scriptsize{}Heads} & {\scriptsize{}Branch} & {\scriptsize{}$\mathcal{L_{\mathrm{\mathit{HEAD-\mathit{MID}}}}}$} & {\scriptsize{}$\mathcal{L_{\mathrm{\mathit{HEAD-\mathit{proposals}}}}}$} & \tabularnewline
\hline 
\textbf{\scriptsize{}$\checked$} & \textbf{\scriptsize{}$\checked$} &  &  &  & {\scriptsize{}35.4}\tabularnewline
\textbf{\scriptsize{}$\checked$} & \textbf{\scriptsize{}$\checked$} &  & \textbf{\scriptsize{}$\checked$} &  & {\scriptsize{}44.5$^{+9.1}$}\tabularnewline
\textbf{\scriptsize{}$\checked$} & \textbf{\scriptsize{}$\checked$} &  & \textbf{\scriptsize{}$\checked$} & \textbf{\scriptsize{}$\checked$} & {\scriptsize{}50.8$^{+6.3}$}\tabularnewline
\textbf{\scriptsize{}$\checked$} & \textbf{\scriptsize{}$\checked$} & \textbf{\scriptsize{}$\checked$} & \textbf{\scriptsize{}$\checked$} & \textbf{\scriptsize{}$\checked$} & \textbf{\scriptsize{}55.3}{\scriptsize{}$^{+4.5}$}\tabularnewline
\hline 
\end{tabular}{\scriptsize\par}
\par\end{centering}
\caption{\label{tab:ablative analysis}Ablative Analysis of EHSOD on VOC07.
We compare the influence of adding CAM branch in RPN, using hybrid-supervised
branches and the effect of adding $L_{\mathrm{\mathit{HEAD-\mathit{proposals}}}}$
in the head. }
\end{table}

\textbf{Ablative Analysis.} We conduct ablation analysis of the proposed
method EHSOD, including the influence of adding CAM branch in RPN,
using hybrid-supervised branches in the head and the effect of adding
$L_{\mathrm{\mathit{HEAD-\mathit{proposals}}}}$. For the hybrid-supervised
head, it can be found that it can boost the performance by 9.1\% mAP
which demonstrates the importance of utilizing the image-level labels
in heads. Adding $L_{\mathrm{\mathit{HEAD-\mathit{proposals}}}}$ can further
improve the mAP by 6.3\%. Our CAM branch in RPN achieves
4.5\% improvements.

\begin{table}
\begin{centering}
\begin{tabular}{c|c|c}
\hline 
{\small{}\# Cascaded modules} & {\small{}mAP$_{50}$} & {\small{}Speed/}\textit{\small{}fps}\tabularnewline
\hline 
{\small{}1} & {\small{}67.8} & {\small{}22.5}\tabularnewline

{\small{}2} & {\small{}70.3$^{+2.5}$} & {\small{}20.1}\tabularnewline

{\small{}3} & {\small{}71.5$^{+3.7}$} & {\small{}18.6}\tabularnewline

{\small{}4} & {\small{}71.6$^{+3.8}$} & {\small{}17.1}\tabularnewline
\hline 
\end{tabular}
\par\end{centering}
\caption{\label{tab: impact of =000023 cascaded modules}Results on the different
number of  proposed Hybrid Supervised Cascade Modules. All models are trained with 50\% of fully-supervised data
and 50\% of weakly-supervised data. All the inference time is tested
on a single V100 GPU. ``3'' is the default
setting of our model.}
\end{table}

\textbf{Impact of Different Number of Cascaded Modules.} We evaluate
the performance of the different number of cascaded modules
of the proposed Hybrid Supervised Cascade Module. The comparison results
are shown in Table \ref{tab: impact of =000023 cascaded modules}.
``Three cascaded modules'' is the default setting of our model.
It can be seen that our hybrid supervised cascade module with three
cascaded modules can significantly improve the performance by 3.7\%
of mAP while only having a runtime overhead with 3.9\textit{fps }out
of 22.5\textit{fps }comparing to the single module. It can be also
found that adding too many cascaded modules (say more than 3) will
not help much.

\section{Conclusion}

We study the hybrid-supervised object detection problem and present
EHSOD, an end-to-end hybrid-supervised object detection system which
can be trained jointly on both fully-annotated data and image-level
data. The performance of the proposed method is comparable to fully-supervised
detection models with only a limited amount of fully annotated-samples, e.g. 37.5
mAP on COCO with 30\% of fully-annotated data. 

\fontsize{9.0pt}{10.0pt} \selectfont
\bibliographystyle{aaai} \bibliography{4084}
\end{document}